\documentclass[accepted]{uai2025} 
                        
\usepackage[american]{babel}

\usepackage{natbib} 
\usepackage{mathtools} 
\usepackage{booktabs} 
\usepackage{tikz} 

\usepackage{algorithm}
\usepackage{algorithmic}
\usepackage{makecell, tabularx, booktabs}
\usepackage{multirow, mathtools, amssymb}
\usepackage{xcolor}
\usepackage{subcaption}
\usepackage{adjustbox}

\title{Multi-Label Bayesian Active Learning with Inter-Label Relationships}

\author[1]{\href{mailto:<yuanyuan.qi@monash.edu>}{Yuanyuan Qi}{}}
\author[1]{\href{mailto:<jueqing.lu@monash.edu>}{Jueqing Lu}{}}
\author[1]{\href{mailto:<xiaohao.yang@monash.edu>}{Xiaohao Yang}{}}
\author[2]{\href{mailto:<joanne.enticott@monash.edu>}{Joanne Enticott}{}}
\author[1]{\href{mailto:<lan.du@monash.edu>}{Lan Du} \thanks{Corresponding author}{}}
\affil[1]{%
    Faculty of Information Technology, Monash University,
    Melbourne, Australia
}
\affil[2]{%
    Faculty of Medicine, Nursing and Health Sciences, Monash University,
    Melbourne, Australia
}
  
\begin{document}
\maketitle

\begin{abstract}
The primary challenge of multi-label active learning, differing it from multi-class active learning, 
lies in assessing the informativeness of an indefinite number of labels while also accounting for the inherited label correlation.
Existing studies either require substantial computational resources to leverage correlations
or fail to fully explore label dependencies. 
Additionally, real-world scenarios often require addressing intrinsic biases stemming 
from imbalanced data distributions. 
In this paper, we propose a new multi-label active learning strategy to address both challenges.
Our method incorporates progressively updated positive and negative correlation matrices to capture
co-occurrence and disjoint relationships within the label space of annotated samples, 
enabling a holistic assessment of uncertainty rather than treating labels as isolated elements. Furthermore, alongside diversity, 
our model employs ensemble pseudo labeling and beta scoring rules to address data imbalances.
Extensive experiments on four realistic datasets demonstrate that our strategy consistently achieves more reliable and superior performance, compared to several established methods. 
\end{abstract}

\section{Introduction}
In recent years, extensive machine learning models and algorithms have been developed to deal with the exponential growth of real-world data. 
However, the significant mismatch between the rapid increase in data and the slow pace of manual data annotation underscores the imperative of active learning (AL) \citep{liu2021influence, xie2022towards}.
Multi-label active learning (MLAL), which considers the co-occurrence of labels and is more aligned with real-world applications, 
has been explored in different domains, including text classification \citep{han2024feature, kang2020active}, 
medical imaging \citep{huang2024uncertainty, simao2023study}, 
remote sensing \citep{mollenbrok2023deep, mollenbrok2023active}, and so on. 
The multi-label task, due to the complexity of label-wise correlation and imbalanced data distribution, 
remains a vital task to comprehensively examine the data, 
thus necessitating the development of effective query strategies \citep{simeoni2020rethinking, dor2020active}. 

To deal with the multi-label issue in active learning, 
earlier approaches usually transform it into multiple binary classification tasks, 
known as binary relevance (BR) which
sums the informativeness evaluated for each individual label to obtain the final acquisition score \citep{zheng2021uncertainty, wang2022attribute}. 
However, these approaches overlook 
the potential correlation of labels, such as their co-occurrence, 
which should be factored into the overall information assessment \citep{zhang2021granular, min2022multi}.
Consequently, the information inherent in the label correlation of the queried samples may not be fully explored.

Some recent works have employed co-occurrence and label correlation matrices to model these inherent label relationships \citep{su2023cost}. 
However, while the positive correlations, 
indicating strong co-occurrence between labels, 
have been included, 
few studies have explored negative correlations
where labels are mutually exclusive and do not appear together.
Moreover, asymmetric label-wise correlations,
where one label frequently appears  with another 
label without a reciprocal relationship,
remains under-explored. 
This also includes the hierarchical structure of the label set, 
where node labels inherently belong to and serve as subsets of their corresponding root labels.
The selection of overlapping labels, due to the hierarchical nature, 
affects the diversity of the strategy, 
consequently, its overall outcome \citep{nakano2020active}.
Furthermore, due to the high imbalance ratios in real-world datasets, 
addressing data imbalance to maintain consistent performance across different datasets 
highlights the critical importance of MLAL tasks \citep{chen2022stable, arens2024rebalancing}.

Considering label co-occurrence and data imbalance, 
we propose a new MLAL framework, 
named multi-label \textbf{C}o\textbf{R}relation-\textbf{A}ware active learning with \textbf{B}eta scoring rules (CRAB) \footnote{Our code is publicly available at \url{https://github.com/qijindou/CRAB}.}
in this paper. 
By incorporating the Beta scoring rules to deal with data imbalance 
and the expected loss reduction framework to select the most informative data instance, 
we introduce dynamic positive and negative correlation matrices to handle the distinct 
and asymmetric label correlation within a Bayesian framework.
This approach demonstrates robust and outstanding performance on four benchmark datasets for multi-label active learning.

\section{Related Work}
Active learning involves selecting the most informative data from the unlabeled pool for annotation, 
thereby reducing the required training data while maintaining comparable performance. 
Two mainstream AL query strategies are uncertainty-based and diversity-based approaches \citep{ren2021a}; 
the former concentrates on informative measurement at the sample-level \citep{abdar2021a}, 
while the latter emphasizes data distribution \citep{kim2023rethink}. 
To quantify the sample uncertainty,
methods such as proper scoring rules \citep{tan2024harnessing}, 
Dirichlet distribution \citep{hemmer2022deal}, and Gaussian Process \citep{shi2021a} 
can be used to estimate the sample informativeness.
By aligning the prior and posterior distributions with model output observations, 
these models can effectively capture the uncertainty.

However, focusing exclusively on uncertainty can introduce bias in sampling 
(i.e., selecting near-identical instances, thus wasting the annotation budget), 
which may lead to sub-optimal performance \citep{prabhu2019sampling}. 
Incorporating diversity into the sampling process offers an alternative approach to 
enhancing generalization \citep{buchert2023toward}. 
\citet{huang2013active} utilized label cardinality inconsistency to exploit uncertainty and integrated it with the diversity-based sampling. 
PLVI-CE leverages average posterior probability discrepancy to measure data diversity and prediction
inconsistency to assess uncertainty, thus enhancing model generalization with limited annotated instances \citep{gu2023plvi}. 
Recently, 
\citet{tan2024harnessing} proposed BESRA which uses the strictly proper scoring rules. 
Its acquisition function combines beta-scoring rules and k-means clustering to enhance diversity,
while the Beta scoring rules also address data imbalance common in multi-label datasets.
Inspired by BESRA, our framework further takes into account label correlation in the acquisition function.

Research in active learning has gradually paid attention to the label correlation in MLAL. 
In light of the specific characteristics of graph data, 
DAMAL incorporates class-label interactions using a graph-based ranking approach, 
where edge weights are defined as the cosine similarity between latent features, 
thus quantifying the graph's informativeness in relation to label correlation \citep{mahapatra2024combining, lu2020multi}. 
To address the uncertainty in feature correlations within standard data, \citet{shi2021a} integrated a Gaussian process with a Bernoulli Mixture model to model correlation through the covariance matrix. 
Correlation matrix-based weighted uncertainty, 
typically derived through co-occurrence or label similarity analysis, 
is commonly used to query the most informative label pairs by capturing the inter-label influence during label selection \citep{gong2021online, su2023cost}. 
\citet{han2024feature} propose a two-stage sample acquisition strategy, called ALMuLa-mix, utilizing inconsistency to capture label correlations with novel features as the first stage and employing the class frequency at the second stage to ensure inter-class diversity. 

Although an increasing number of studies recognize the importance of correlation during data acquisition, existing approaches are often resource-intensive, requiring additional training for interrelation modeling, or struggle to maintain performance under data imbalance.
Our approach effectively samples the representative data in a correlation-aware manner while maintaining consistent performance, even with highly imbalanced datasets.

\section{Correlation-aware Multi-label Active Learning}
\begin{figure*}[t]
    \centering
    \includegraphics[width=0.98\textwidth]{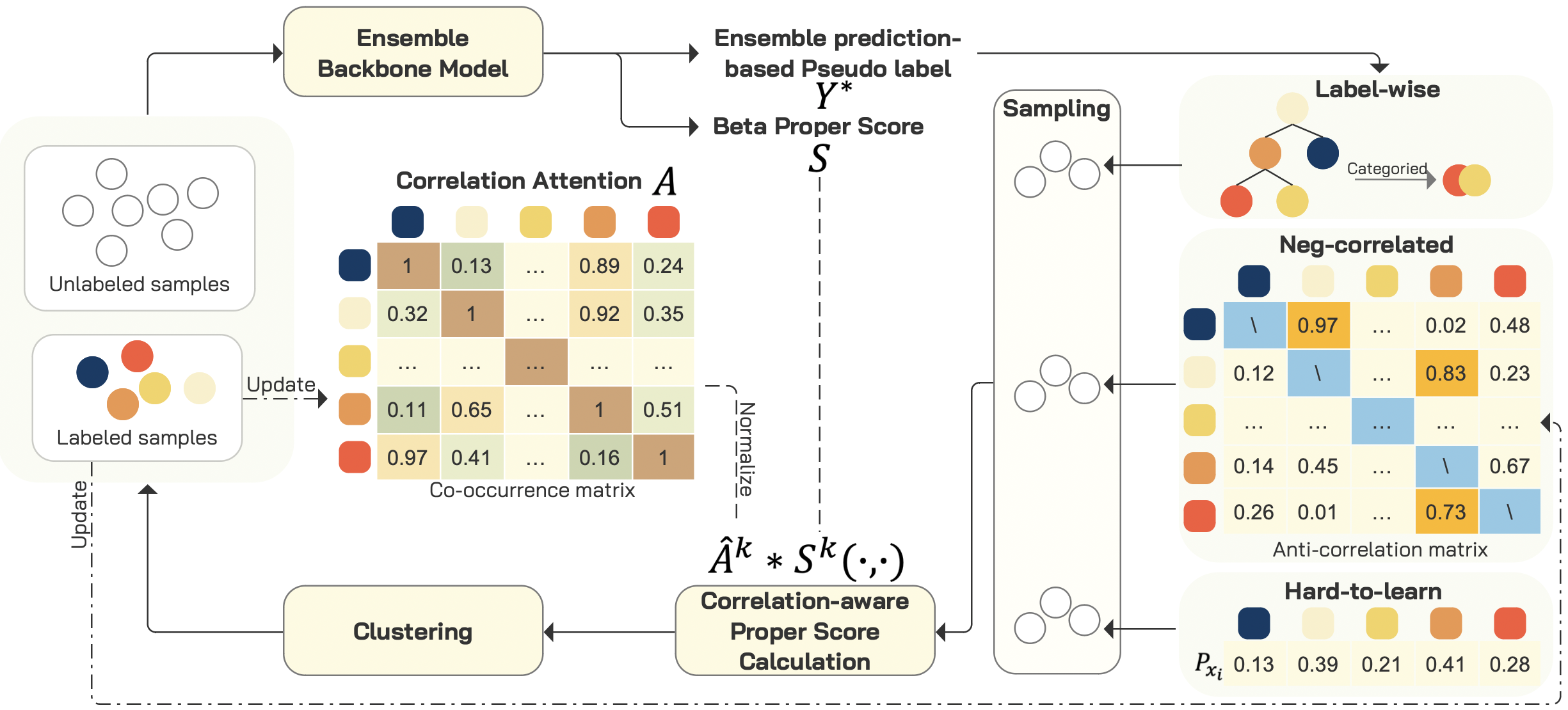}
    \caption{Overview of the CRAB framework. 
    It is trained using an ensemble method that generates pseudo labels based on predictions and calculates the beta proper score. 
    Firstly, positive and negative correlation matrices are updated with newly sampled instances. 
    Then, utilizing pseudo labels, our model samples data from three categories: label-wise, negatively correlated, and hard-to-learn samples. 
    Last, our model calculates correlation-aware proper score and subsequently clusters and labels the selected data based on this score.
    \label{fig:flowchart}}
\end{figure*}
Without loss of generality, suppose $L=\{X,Y\}$, 
$U=\{X\}$ represent the initial collection of training set and unlabeled data samples, where
$|U| \gg |L|$; and $y_i \in \{-1, +1\}^k$ represents the label of the $i_{th}$ example in the $k$ label space. $K$ denotes the total number of labels in the space.
Firstly, our model generates the two-dimensional correlation matrix, 
including both positive correlation and negative correlation,
based on the iteratively updated labeled dataset $L$.
Then, given a model parameterized by $\theta\in\Theta$, 
the probability of label $y$ of a data instance $x$ is $P(y|\theta, x)$.
We are able to derive the pseudo label as $y^*$ based on $\int_\theta P(y|\theta,x)P(\theta)d\theta$,
where the integration can be approximated by Monte Carlo via ensemble.
Considering the model's learning capability of different categories of data,
our model refines the sampling pool into a more preventative subset based on the pseudo labels.
And considering the influence of label correlation in  quantifying the informativeness of sample, we propose a variation of the beta scoring rule used in \citet{tan2024harnessing}. Its key idea is introduced in Section~\ref{sec:caq}.
Finally, the clustering approach assures the diversity in sampling. 
Fig.~\ref{fig:flowchart} illustrates the overall flowchart of our proposed framework. 

\subsection{Preliminaries}\label{sec:beta}
To address the multi-label active learning problem, most studies decompose it into multiple binary classification tasks, aggregating individual label scores instead of assessing the entire instance holistically. This approach is formulated in Eq.~\eqref{eq0}, where $S_{BR}$ represents a scoring function that measures the informativeness (e.g., uncertainty score) of individual samples, and $S_{BR}^k$ denotes the score with respect to each label.
\begin{align}
    S_{BR}(p,y)  =  \sum^K_{k=1}S^k_{BR}(p,y^k)  &= \sum^K_{k=1} \mathcal{L}(y^k|p) \label{eq0}
\end{align}
Monte Carlo estimation  offers a probabilistic framework for computing acquisition scores by incorporating randomness to account for variability. 
Monte Carlo-based error reduction estimation \citep{roy2001toward} utilizes the Monte Carlo approach to approximate the expected reduction in error resulting from the labeling of a given sample. 
However, while Monte Carlo-based methods estimate expected improvement in model performance, 
they do not assess the quality of probabilistic outputs.
Rather than estimating error, proper scoring rules provide a summary measure of predictive probability, computing the positive-oriented rewards (i.e., utilities) that a classifier seeks to maximize \citep{gneiting2007strictly}.
Eq.~\eqref{eq1} and \eqref{eq2} show the core concept of expected increase in score when querying. 
\begin{align}
    Q(L) & =  \mathbb{E}_{P_{(x)}}\mathbb{E}_{P(\theta|L)}\big [\mathbb{E}_{P(y|\theta,x)} \notag\\
              & \quad\quad [S(P(\cdot|x,\theta),y)-S(P(\cdot|x),y)]\big ] \label{eq1} 
\end{align}
\begin{align}
    \Delta Q(x|L)  & =  Q_L - \mathbb{E}_{P_{(y|L,x)}}[Q_{L+\{x, y\}}] \notag\\
                        & =  \mathbb{E}_{P_{(y|L,x)}} \big[ \mathbb{E}_{P(x')P(y'|L,(x,y),x')} \label{eq2}\\
                        & \quad\quad [S(P(\cdot|L,(x,y),x'),y')-S(P(\cdot|L,x'),y')] \big]
                        \nonumber
\end{align}
$S$ denotes the scoring function that evaluates predictive probability distribution on an event, i.e., predicting $y'$, given $x'$. 
$Q(L)$ represents the mean proper scoring rule of the predictive probabilities obtained using Bayesian estimation based on the current labeled dataset $L$.
The term $\Delta Q(x|L)$ denotes the increment in the score resulting from acquiring the label of a sample $x$, drawn from the unlabeled data pool $U$. 
And $x$ is the point to be acquired, $x'$ denotes the selected unlabeled anchor point for assessment.
The value of $\Delta Q(x|L)$ is then used to select sample leading to a large increment in the score or reward. 
Since the label of unlabeled points $x$ and $x'$ is unknown, we derive $P(y'|L, (x,y),x')$ by calculating the posterior distribution of the ensemble models using Eq.~\eqref{eq3}. 
\begin{align}
    P(y'|L,(x,y),x')   &= \mathop{\sum}\limits_{\theta\in\Theta^E}P(y'|\theta,x')P(\theta|L,(x,y)) \label{eq3}
\end{align}
\begin{align}
    P(\theta|L,(x,y))  &\approx \frac{P(\theta|L)P(y|\theta,x)}{\sum_{\theta\in\Theta^E}P(\theta|L)P(y|\theta,x)} \label{eq4}
\end{align}
Beta family \citep{buja2005loss}, which generalizes the logarithmic score and the brier score, or the other desired cost-weighted scoring rule, is able to address the issue of imbalanced label distribution in multi-label learning.
The equation below illustrates the proper scoring rules $\mathcal{L}$ of a predictive distribution $p$ given the expected value $y^k$, where $y^k$ represents the label for class $k$. $\mathcal{L}(1|p)$ and $\mathcal{L}(0|p)$ represent the partial losses when $p$ is been classified as 1 and 0, respectively.
\begin{align}
    S^k_{BR}(p,y^k) \quad = \quad y^k \mathcal{L}(1|p) + (1-y^k)\mathcal{L}(0|p) 
\end{align}
\begin{align}
    \mathcal{L}(1|p) &=  \mathcal{L}_1(1-p)  =  \textstyle \int^1_pp^{\alpha-1}(1-p)^{\beta}dp \\
    \mathcal{L}(0|p) &=  \mathcal{L}_0(p)  =  \textstyle \int^p_0p^{\alpha}(1-p)^{\beta-1}dp 
\end{align}
By leveraging $I_x(\alpha,\beta)$, the Incomplete Beta Function, the closed form of the Beta distribution is obtained for $\alpha, \beta > 0$. When $\alpha=\beta=0$, the scoring becomes log-loss, and when $\alpha=\beta=1$, the scoring rule will transform to squared error losses. By adjusting the value of $\alpha$ and $\beta$, our model can effectively handle scenarios with diverse data distributions. In our research, we employ the greedy search result of BESRA as the parameter for scoring, where the $\alpha=0.1, \beta=3$ \citep{tan2024harnessing}.

\subsection{Correlation Matrix Construction}
In multi-label scenario, a single sample often has more than one label, and label with relatively small semantic distances frequently appear simultaneously. 
To leverage inherent relationships within the label space to assist in acquisition process, 
our framework maintains two dynamic matrices: a co-occurrence matrix representing positive correlations, and a anti-correlation matrix representing negative correlations between labels. Both matrices are updated after each acquisition iteration with newly annotated instances. By discovering the pattern of the occurrence between labels, we aim to quantify the informativeness considering influence of correlation, and maintain the diversity while take into account the imbalanced data distribution.

\textbf{Positive correlation matrix}: 
The positive correlation matrix $A$ is constructed based on the label-wise dependence.
$A$ is a $K\times K$ two dimensional matrix, where each element $A(m,n)$ quantifies the dependency of the presence of label $m$ on label $n$. This dependency is formally computed using Eq.~\eqref{eq8}.
Specifically, $\sum_{i=1}^L N(y_i^m=+1, y_i^n=+1)$ refers to the count of labeled instances 
in which both labels $m$ and $n$ appear simultaneously.
$P(y^{m}|y^{n})$ gives the likelihood of label $m$ occurring when label $n$ is present. 
When $m=n$, $A(m,n)$ equals 1. 
The value of $A(m,n)$ reflects the probability of one label's existence conditioned on  another. 
\begin{align}
    A(m,n) \quad &= \quad P(y^{m}|y^{n}), \quad \text{where } m\neq n \notag \\
                 &= \quad \frac{\sum_{i=1}^L N(y_i^m=+1, y_i^n=+1)}{\sum_{i=1}^L N(y_i^n=+1)} 
    \label{eq8}
\end{align}
The positive correlation matrix is constructed to characterize the pattern of the co-occurrence between labels and capture the asymmetric correlation between labels, including hierarchical relationships.

\textbf{Negative correlation matrix}: 
Despite the positive correlations, negative correlations between labels have rarely been addressed in previous research.
However, in real-world scenarios, negative correlations are instrumental in  enabling the model differentiate between mutually exclusive classes and contribute to more accurate decisions by clarifying the model's decision boundaries \citep{yang2024not}. 

To effectively model these negative correlations, we construct an updated anti-correlation matrix, $NegA$. 
Maintaining the same format as the positive correlation matrix $A$, $NegA$ is a $K\times K$ two dimensional matrix, where each element $Neg(m,n)$ quantifies the confidence in the absence of label $m$ given the presence of label $n$, as defined in Eq.~\eqref{eq9}.
Specifically, $\sum_{i=1}^L N(y_i^m=-1, y_i^n=+1)$ represents the count of instances where labels $m$ and $n$ do not co-occur.  
This asymmetry allows the matrix to capture the nuanced conditional negative relationships, and provide a new perspective towards the label dependencies.
\begin{align}
    NegA(m,n) \quad &= \quad P(\overline{y^m}|y^n), \quad \text{where } m\neq n \notag\\
                     &= \quad \frac{\sum_{i=1}^L N(y_i^m=-1, y_i^n=+1)}{\sum_{i=1}^L N(y_i^n=+1)}
    \label{eq9}
\end{align}
\subsection{Correlation-based Sampling}\label{sec:corr-sample}
Refining the unlabeled pool to ensure that selected instances concentrate on specific representative criteria is a common strategy in active learning \citep{kang2020active}. 
However, current research predominantly based on informativeness analysis, neglecting the critical role of data correlation in MLAL.
To address this limitation and provide more representative and evenly distribution samples for continuous process,
our model refines the unlabeled pool from three perspectives based on the correlation properties to generate a subset to be used in acquisition. And the pseudo label $y^*$, obtained by averaging the prediction result of ensemble models through Eq.~\eqref{eq10}, is used for the following correlation-based sampling.
\begin{align}
    y^* &= \mathbb{I}[P(y \mid x, L)>0.5] \label{eq10}
\end{align}
\begin{align}
    P(y \mid x, L) &= \textstyle \int_{\theta} P(y|x,\theta) p(\theta \mid L) \notag \\
    &\approx \textstyle \sum^E_{e=1} P(y|x,\theta_e)/E\label{eq}
\end{align}
\subsubsection{Label-wise sampling}
In multi-label scenarios, labels often exhibit asymmetric correlations, where the present of one label, $m$, is highly correlated with another label, $n$, but not vice versa. Hierarchical structures within labels are a common example of this kind of relationship. To illustrate the impact on performance, we can consider hierarchical data:
when asymmetric correlations exist, the selection of root-node labels often inevitably overlaps with that of corresponding leaf-node labels, while rarely occurring independently, which reduces the representativeness of the selected root labels \citep{nakano2020active}.

To address this, our strategy introduces a new mechanism for the label-wise selection. 
If one label $n$ is highly dependent on the presence of another $m$, 
while the reverse is not necessarily true—otherwise, 
they would effectively be considered the same label in most cases—and their correlation exceeds a predefined threshold, $\sigma$, set as the standard deviation of a two-tailed normal distribution, we classify the label pair as asymmetrically correlated. To improve label-wise sampling, we then refine the label space of instances that contain both labels $m$ and $n$ by removing label $n$, 
ensuring that sampling prioritizes the most independent label, $m$.
The model then performs evenly sampling based on the refined pseudo labels, ensuring a more balanced sampling pool.

\subsubsection{Negative-correlated label sampling}
In multi-label learning, it is essential for the model to respect the exclusivity of certain labels to ensure accurate predictions \citep{huang2017multi}.
When mutually exclusive labels, such as those that should not logically co-occur, are predicted together, it is often an indication of model bias or misguided learning \citep{huang2021local, perales2020negative}. 
This misalignment can reduce model's effectiveness. 
As the second subset for concentrated sampling, we select instances with negatively correlated pseudo labels that are not expected to co-occur in the label space.

To formalize this, we consider a pair of labels as mutually exclusive when the negative correlation coefficient $NegA(m,n)$ exceeds a predefined threshold, set as $2\sigma$, where $\sigma$ represents the standard deviation.
Based on predictions with pseudo labels, our model selects samples with the predicted labels that are unlikely to co-occur, according to the negative correlation matrix, as those samples are at high risk of incorrect predictions. These selected samples are then added to the refined subset of the unlabeled pool, ensuring that the model better accounts for negative correlations.

\subsubsection{Hard-to-learn label sampling}
The third set, which our model uses to further expand the subset, 
consists of hard-to-learn samples. 
These samples are typically characterized by low confidence and low variability, 
indicating instances where the model has difficulty making accurate predictions. 
Such samples often contain ambiguous or noisy features or lie near decision boundaries \citep{chang2017active, yang2020rethinking}. 
In this study, we define samples without any predicted pseudo labels as hard-to-learn samples. Specifically, if the pseudo labels obtained through Eq.~\eqref{eq10} for all classes of a given instance falls below the threshold, 0.5, the classifier cannot make any prediction, thus the sample is classified as hard to learn. 
To improve performance on these challenging instances while maintaining diversity in the sampling process, our model dynamically adjusts the sample size using a polynomial decay function, enabling more focused learning on difficult cases over time.
\begin{algorithm} [!t]
    \caption{CRAB Update Strategy for MLAL}
    \label{alg:algorithm}
    \textbf{Input}: Labeled pool: $L$; Unlabeled pool: $U$; Model: $\Theta^E={\theta_1,...,\theta_E}$; Query size: $N$; Per-label query size: $N$; Hard to learn query size: $Z$.\\
    \textbf{Output}: Updated labeled and unlabeled pool.\\
    \begin{algorithmic}[1]
        \FOR{$\theta_e \in \Theta^E$}
        \STATE Get the prediction $P(y|\theta_e,x)$ and corresponding Beta score $S_{BR}$
        \ENDFOR
        \STATE Update correlation matrix $A$ and $NegA$ with $L$ via Eq.~\eqref{eq8} and Eq.~\eqref{eq9}
        \STATE Get the pseudo label $Y^*$ via Eq.~\eqref{eq10}
        \STATE Select and add $N$ label-wise samples per label to $U^*$
        \STATE Select and add $N$ negative-correlated samples with $NegA$ to $U^*$
        \STATE Select and add $Z$ hard samples to $U^*$
        \FOR{$u \in U^*$}
        \STATE Get the attention beta score, $S_{AB}$, via Eq.~\eqref{eq11}. Update to $Q(x|L,x')$ with Eq.~\eqref{eq2}
        \ENDFOR
        \STATE $L^+$=$k$-Means Centers($Q(x|L,x')$, $N$)
        \STATE $L \leftarrow L+L^+$; $U \leftarrow U-L^+$
        \STATE \textbf{return} $L,U$
    \end{algorithmic}
\end{algorithm}

\subsection{Correlation-aware Querying}
\label{sec:caq}
With the correlation-based sampling strategy described in section~ \ref{sec:corr-sample}, our model obtains a refined subset of the original unlabeled data pool.
Then, our model calculates correlation-aware beta scores for these samples in the selected subset, and use those scores to cluster the samples for acquisition. 
This score computation method is inspired by the attention mechanism introduced in transformer models \citep{vaswani2017attention}, which is defined as follows. 
\begin{align}
    S_{AB}(f_L(x),y)     &=  \sum^K_{m=1} \hat{A}(m,:) S_{BR}^m(f_L(x),y^m) \label{eq11}
\end{align}
\begin{align}
    \hat{A}(m,n)&=  \text{norm}(A(m,n)), \quad \text{where } m \neq n \notag \\
                &=  \frac{A(m,n)}{\gamma \cdot \text{max}(A(:,n))} \label{eq13}
\end{align}
Using Eq.~\eqref{eq11}, we score each prediction, where $S_{AB}$ incorporates the influence of other labels' scores through the attention coefficient $\hat{A}$ as the final score, accounting the correlation. 
Additionally, we introduce $\gamma$, a normalization parameter set to 2, to prevent over-estimating correlated uncertainty while preserving the original significance of each label's score. 
This approach allows our model consider the impact of neighboring labels on informativeness, and the refined unlabeled pool enhances computational efficiency and deepens the analysis of label correlations. 
Algorithm~\ref{alg:algorithm} details the procedure for one iteration of our framework.

\section{Experiments}
We collected four benchmark multi-label text datasets to analyze the performance and robustness of our framework \citep{kement2023exploration}. 
Those datasets include: 
RCV1 \citep{lewis2004rcv1},
UKLEX \citep{chalkidis2022improved},
EURLEX \citep{chalkidis2021multieurlex},
and MIMIC3 \citep{johnson2016mimic}.
RCV1  is a news articles dataset from Reuters; 
UKLEX is a collection of legal documents sourced from various categories within UK law; 
EURLEX is a set of descriptors from European legal information thesaurus extracted from the European Union's legal database; 
and MIMIC3 is a set of de-identified health records for medical diagnosis. 
Following the method by \citet{charte2015addressing}, we used mean imbalance ratio (MeanIR) to create synthetic datasets with varying imbalance ratios based on the modified RCV1 dataset, reduce the label size of RCV1 to ten by selecting the most frequently occurring labels, enabling an evaluation of the model's performance across different degrees of imbalance.
Table~\ref{tab:data} and Table~\ref{tab:data_ir} offer a detailed summary of these four datasets. 
We also introduced a new metric, termed \textbf{CorrAvg}, 
defined as $\sum_{m=1}^K\sum_{n=1}^K$ $A(m,n)/(K\times K), m\neq n$, to quantify the degree of inter-correlation within label set. 

\subsection{Implemetation}
We used Neural-Classifier \citep{liu2019neuralclassifier}, implemented in Pytorch \citep{paszke2019pytorch}, as the code base. 
In our study, we exployed three mainstream models, TextCNN \citep{zhang2017sensitivity}, TextRNN \citep{liu2016recurrent}, and DistilBERT \citep{sanh2019distilbert}, as the backbone classifiers. 
To enhance efficiency and performance,
we applied the cold start strategy \citep{zhu2019addressing} with random initialization at the beginning of each active learning iteration, a method known for its applicability to real-world scenarios \citep{frankle2018lottery}. 
All experiments were conducted on a single RTX3090 GPU. 
Following the setting of \citet{tan2024harnessing}, 
the maximum sequence length for the text data was set to 256, 
with each training iteration consisting of 80 epochs. 
The initial training set and validation set sizes are set to 100 and 1000, respectively, and are sampled from the training set.
We implemented an early stopping criterion with the patience of 30 epochs to prevent the model from falling into local optima or overfitting \citep{du2019gradient, ying2019overview}. 
AdamW was used as the optimizer \citep{loshchilovdecoupled}, 
with the learning rate tailored for each model: 5e-2 for TextCNN and TextRNN, and 5e-5 for DistilBERT.
The hard-to-learn query size was set to 300 for benchmark datasets and 200 for synthetic datasets, while the per-label query size was set to 50 for RCV1 and 100 for other datasets, due to differences in label space size.
\begin{table}
    \adjustbox{max width=\linewidth}{%
    \centering
    \begin{tabular}{lcccc}
        \toprule
        \bfseries Dataset & \makecell[l]{\bfseries \#Document\\ \bfseries Train/Test} & \makecell[l]{\bfseries \#Vocab./\\ \bfseries \#Label} & \makecell[l]{\bfseries \#MeanIR\\ \bfseries Train/Test} & \makecell[l]{\bfseries \#CorrAvg\\ \bfseries Train/Test} \\
        \midrule
        RCV1 & 24,891/6223 & 104,619/102 & 402/197 & 0.137/0.137\\
        UKLEX & 20,000/8500 & 63,157/18 & 7/6 & 0.026/0.024\\
        EURLEX & 55,000/5000 & 160,211/21 & 16/15 & 0.131/0.147 \\
        MIMIC & 29,999/10000 & 137,678/19 & 127/101 & 0.321/0.320 \\
        \bottomrule
    \end{tabular}}
    \caption{Benchmark datasets with corresponding imbalance level and correlation level statistics.}
    \label{tab:data}
\end{table}
\begin{table}
    \adjustbox{max width=\linewidth}{%
    \centering
    \begin{tabular}{lcccc}
        \toprule
        \textbf{Dataset} & \makecell[l]{\textbf{\#Document}\\ \textbf{Train/Test}} & \makecell[l]{\textbf{\#Vocab./}\\ \textbf{\#Label}} & \makecell[l]{\textbf{\#MeanIR}\\ \textbf{Train/Test}} & \makecell[l]{\textbf{\#CorrAvg}\\ \textbf{Train/Test}} \\
        \midrule
        RCV1-T10-5 & 1,200/600 & 25,254/10 & 5/10 & 0.133/0.138\\
        RCV1-T10-10 & 1,200/600 & 25,289/10 & 10/10 & 0.135/0.138\\
        RCV1-T10-20 & 1,200/600 & 24,170/10 & 20/10 & 0.137/0.138\\
        RCV1-T10-50 & 1,200/600 & 25,280/10 & 50/10 & 0.142/0.138\\
        \bottomrule
    \end{tabular}}
    \caption{Synthetic datasets with corresponding imbalance level and correlation level statistics.}
    \label{tab:data_ir}
\end{table}

\subsection{Baselines}
To conduct a comparative performance analysis, 
we adopted five state-of-the-art MLAL methods as baselines, 
including random sampling. 
Each baseline uses the same query parameters and backbone classifier to maintain consistency across experiments. 
Specifically, \textbf{MMC} \citep{yang2009effective} applies maximal confidence to selecting data that induces the largest reduction in expected model loss. 
\textbf{AUDI} \citep{huang2013active} explores uncertainty and diversity in both instance and label spaces through label ranking and threshold learning. 
\textbf{ADAPTIVE}\citep{li2013active} integrates max-margin prediction uncertainty with label cardinality inconsistency to assess the unified informativeness of multi-label instances. 
\textbf{BESRA} \citep{tan2024harnessing} utilizes the beta scoring rules within an expected loss reduction framework to evaluate informativeness and employs vector representations to maintain diversity. 
\textbf{CMAL} \citep{yu2020cmal} leverages global label correlation matrix and label space sparsity with uncertainty to query the most informative example-label pairs.

\begin{figure}
    \centering
    \includegraphics[width=0.95\linewidth]{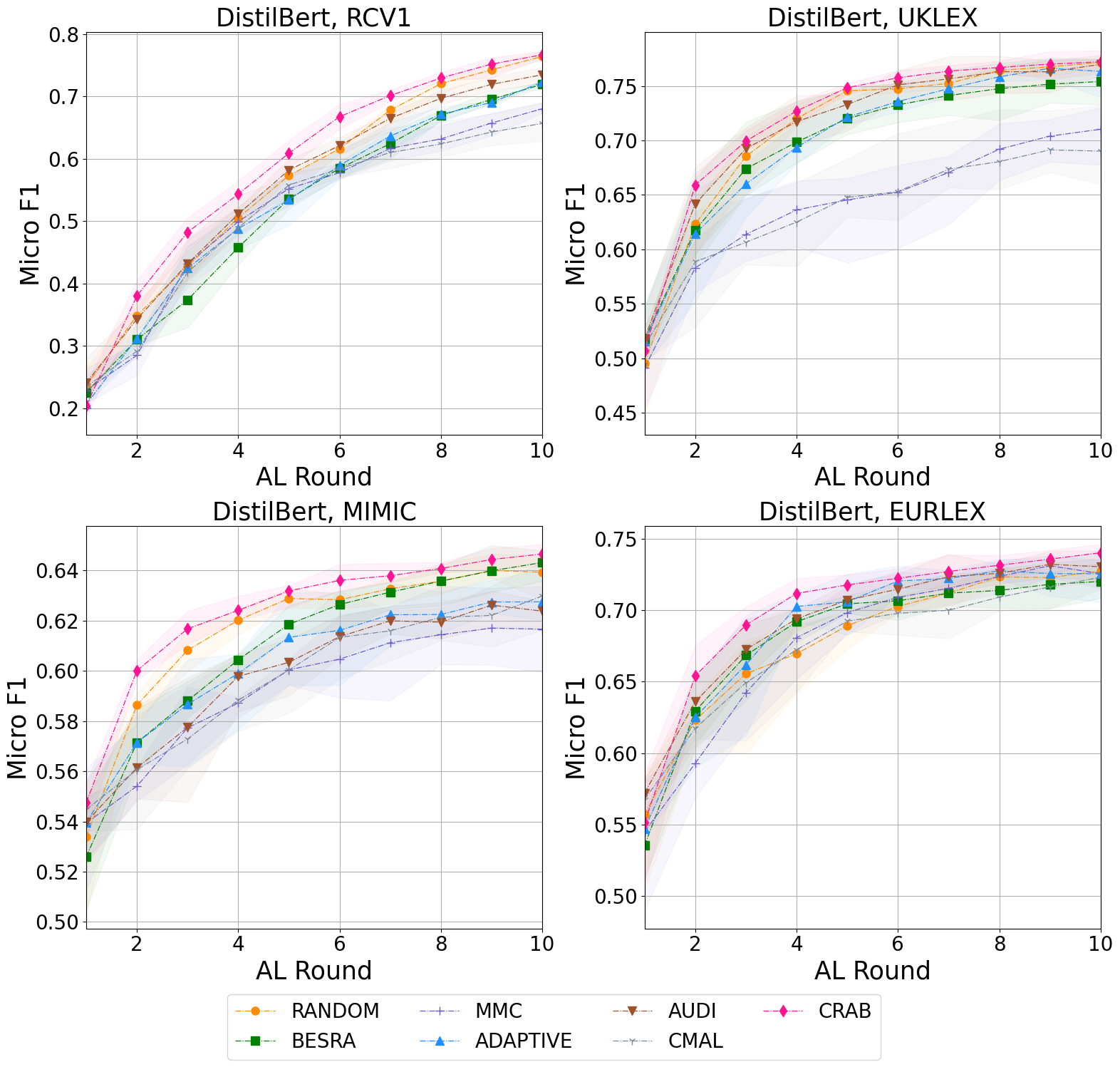}
    \caption{Averaged micro-F1 score on DistilBERT, averaged results with 5 random seeds.}
    \label{fig:bert}
\end{figure}

\subsection{Results}
\begin{figure}
    \centering
    \includegraphics[width=0.95\linewidth]{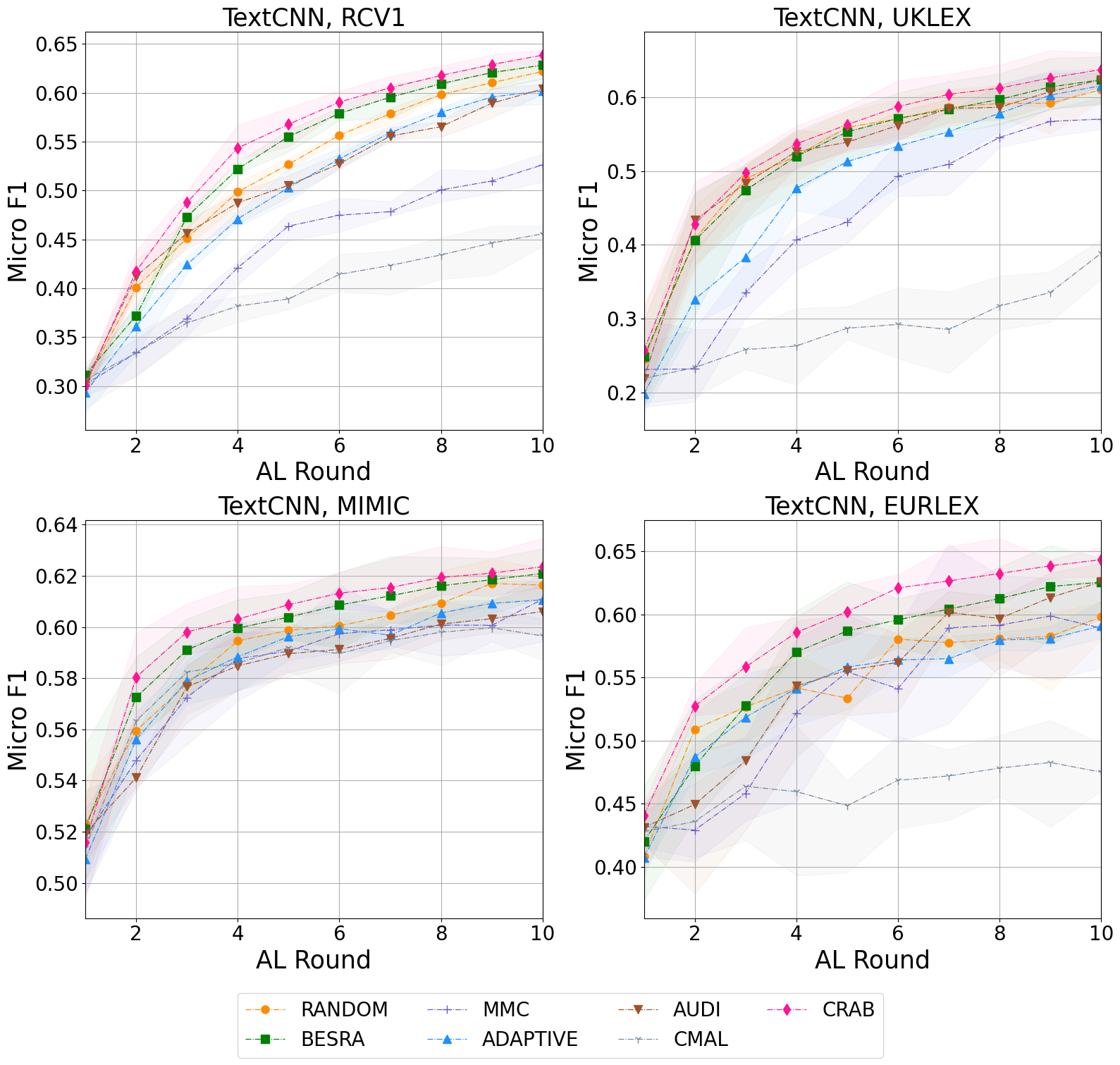}
    \caption{Averaged micro-F1 score on TextCNN, averaged results with 5 random seeds.}
    \label{fig:textcnn}
\end{figure}

\begin{figure}
    \centering
    \includegraphics[width=0.95\linewidth]{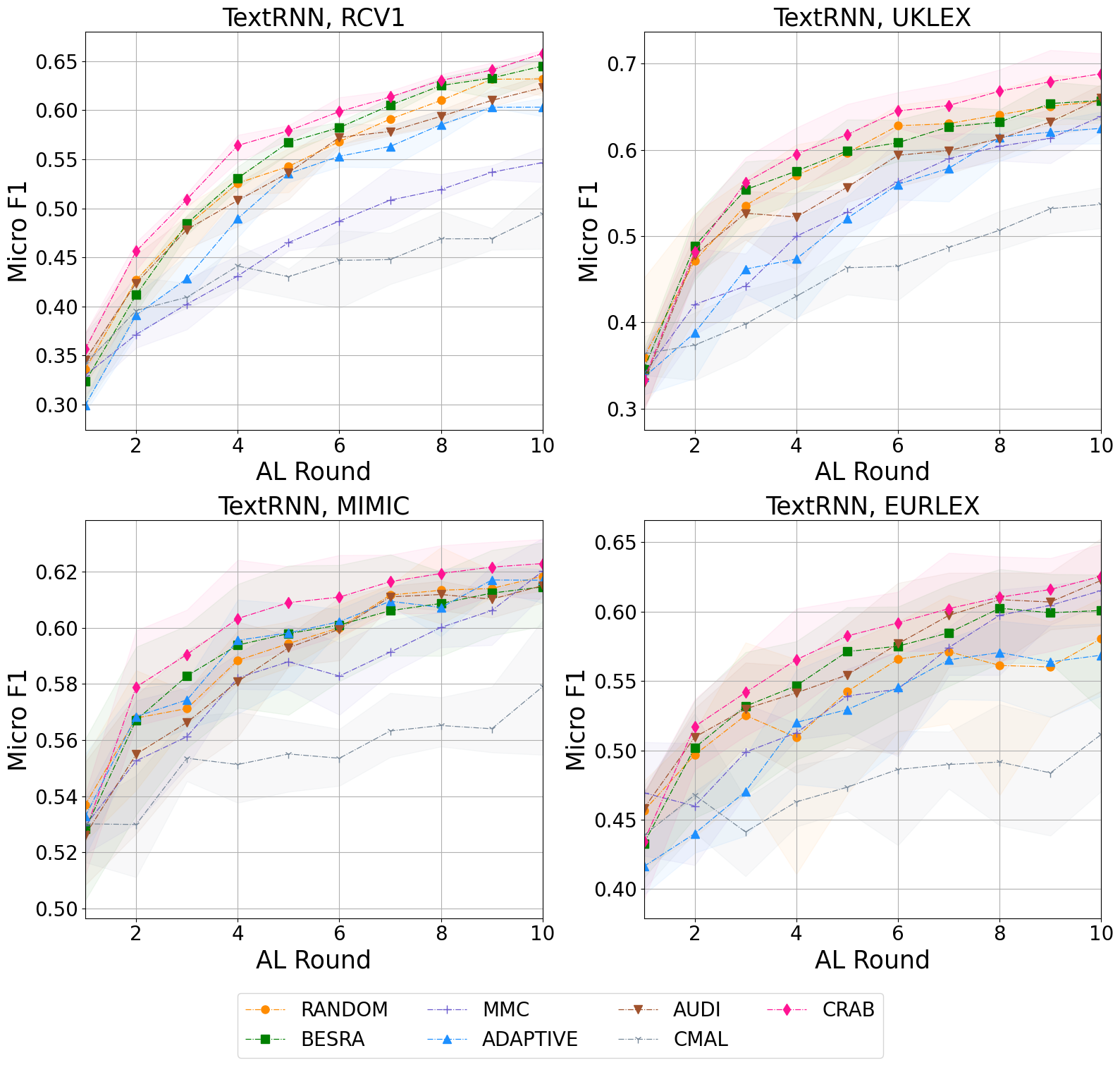}
    \caption{Averaged micro-F1 score on TextRNN, averaged results with 5 random seeds.}
    \label{fig:textrnn}
\end{figure}

\begin{figure}
    \centering
    \includegraphics[width=0.95\linewidth]{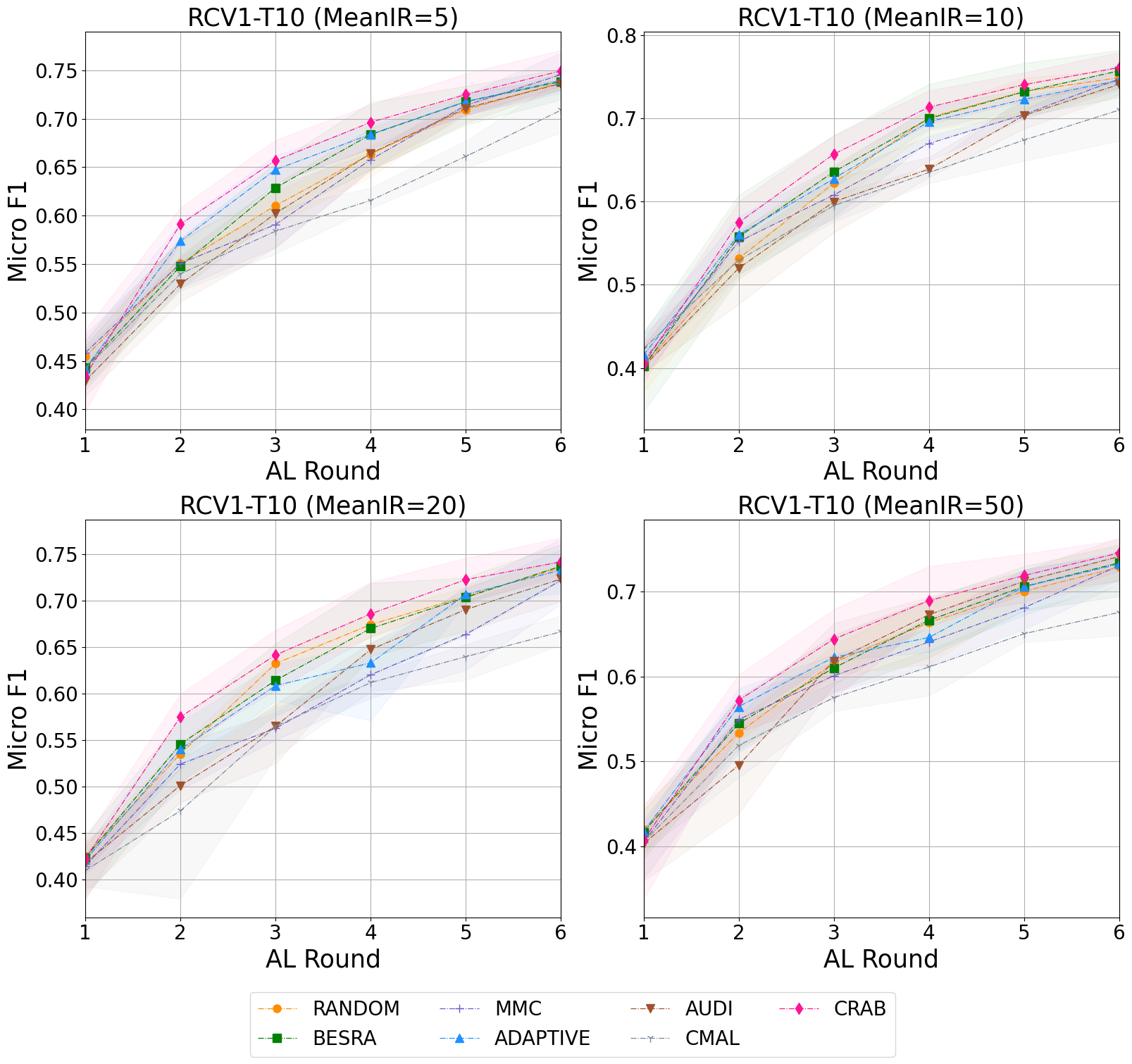}
    \caption{Averaged micro-F1 score on synthetic dataset using TextCNN, averaged results with 5 random seeds.}
    \label{fig:ir}
\end{figure}

Figures~\ref{fig:bert}--\ref{fig:ir} present the quantitative performance of our proposed framework. 
Figure~\ref{fig:bert} compares the performance of CRAB and baseline methods across four datasets using DistilBERT. 
Figure~\ref{fig:textcnn} and Figure~\ref{fig:textrnn} present supplementary performance comparisons based on TextCNN and TextRNN.
Following \citet{tan2024harnessing},
we obtained the predictive distribution by training five ensemble models independently,
each initialized with the same parameters for every AL iteration. 
The micro-F1 results show that our proposed model, CRAB, has consistently outperforms other AL methods across different text domains and network structures. 
Additionally, CRAB demonstrates robust performance on datasets with varying degrees of correlation, particularly compared with BESRA,
suggesting that our strategy effectively models correlation during data selection. 

Among the baseline methods, BESRA achieves strong results and shows relatively robustness on different datasets. However, its performance on the highly correlated MIMIC dataset is less stable, likely due to the absence of correlation consideration. AUDI, which incorporates both uncertainty and diversity at both data and instance level, presents notable performance on three of the datasets, RCV1, UKLEX, and EURLEX, but struggles on MIMIC. Adaptive and MMC yield similar results over four datasets, as both utilize the max-margin as the selection criterion. 
Although CMAL considers global label correlation, it only performs optimally on the highly correlated MIMIC dataset and does not maintain stable performance on all datasets. 
To examine robustness of the model on imbalanced datasets, 
we conducted comparative experiments on synthetic datasets, with results shown in Figure~\ref{fig:ir}. 
CRAB maintains superior performance across synthetic datasets with different MeanIR values, demonstrating its capability to handle imbalanced datasets.

\begin{figure}
    \centering
    \includegraphics[width=0.95\linewidth]{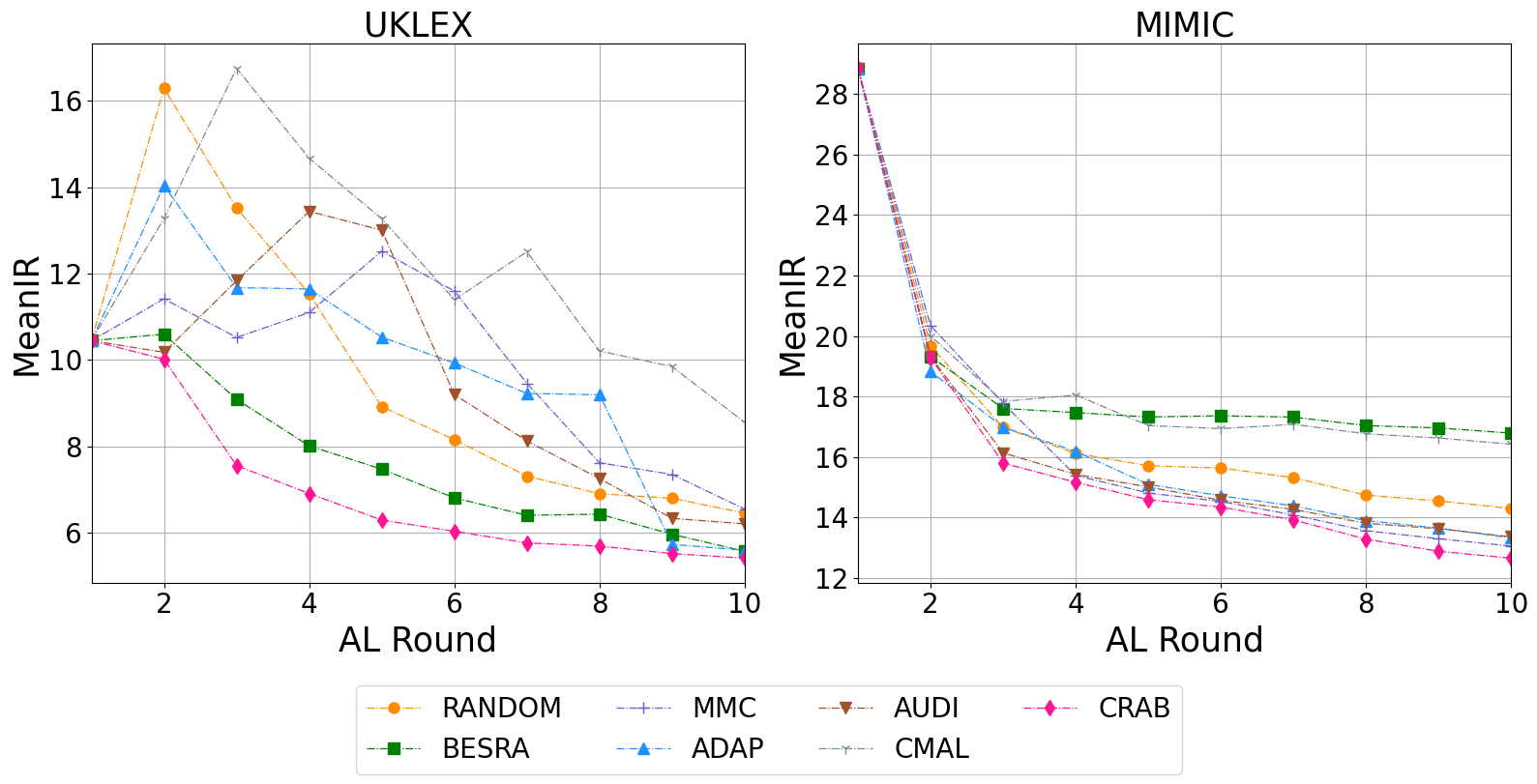}
    \caption{The averaged MeanIR of selected samples, averaged the results with 5 random seeds.}
    \label{fig:meanir}
\end{figure}

\begin{figure}
    \centering
    \includegraphics[width=0.95\linewidth]{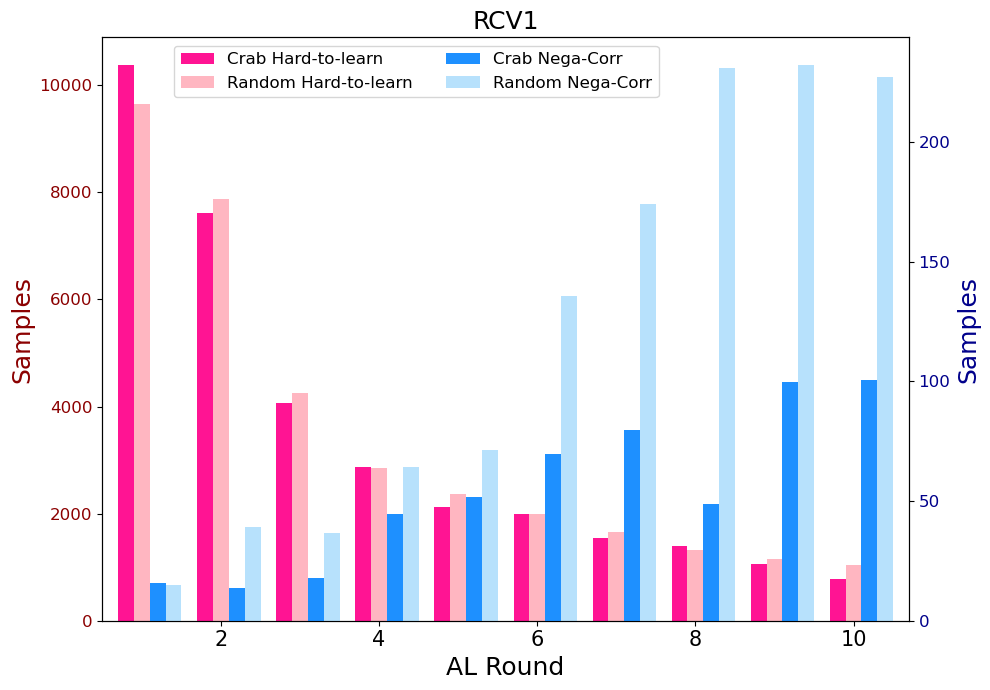}
    \caption{Trend of hard-to-learn and negative-correlated data, with the red bar axis on left and the blue bar axis on right, averaged the results with 5 random seeds.}
    \label{fig:trend}
\end{figure}

To further investigate the effectiveness of our model, 
Figures~\ref{fig:meanir} and \ref{fig:trend} present a qualitative analysis of CRAB. 
Figure~\ref{fig:meanir} shows the MeanIR of the selected samples across AL iterations. 
Since MeanIR indicates imbalance, with lower values reflecting a more even data distribution, 
we observe that CRAB demonstrates a more balanced sample selection compared to other baseline models, 
which underscores its capacity to address data imbalance effectively.
Figure~\ref{fig:trend} illustrates the trend of two categories of data within the unlabeled data pool: hard-to-learn data and negatively correlated data.
Unlike random selection, CRAB strategically selects data that enhances model learning,
thereby reducing misclassification of negatively correlated data. 
Additionally, CRAB improves performance on hard-to-learn data,
helping the model becomes more robust and accurate. 
This targeted data selection contributes to a more balanced and adaptive learning process, 
ultimately leading to improved generaliztion across diverse data types.


\subsection{Parameter sensitivity analysis}
\begin{figure}
    \centering
    \begin{subfigure}{0.49\linewidth}
        \centering
        \includegraphics[width=\linewidth]{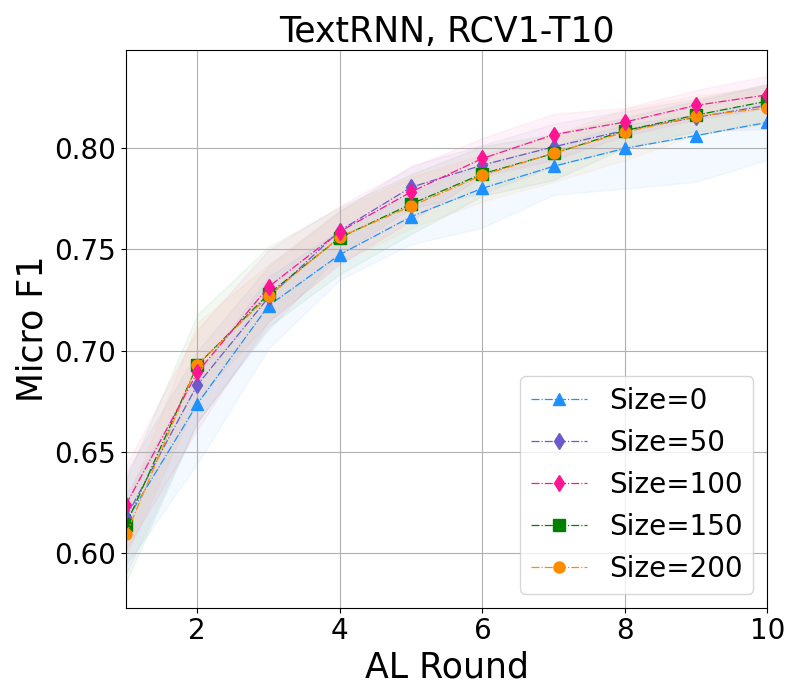}
        \caption{}
        \label{fig:empty}
    \end{subfigure}
    \hfill
    \begin{subfigure}{0.49\linewidth}
        \centering
        \includegraphics[width=\linewidth]{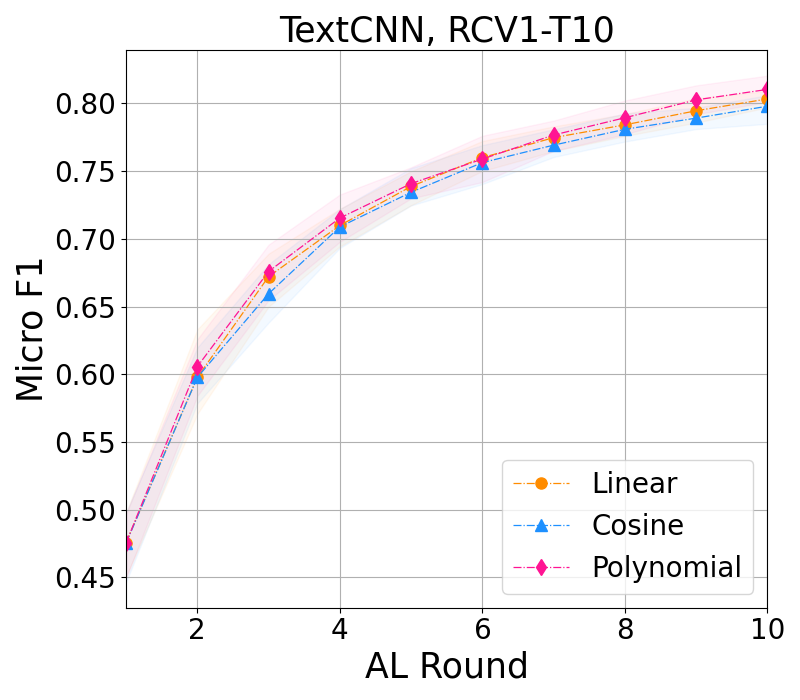}
        \caption{}
        \label{fig:decay}
    \end{subfigure}
    \caption{(a) Performance for different size of hard-to-learn samples. (b) Performance for different decay functions of the hard-to-learn samples.}
    \label{fig:empty_decay}
\end{figure}

To validate the effectiveness and generalizability of CRAB, we conduct two experiments to analyze its performance under different parameter settings. Figure~\ref{fig:empty} demonstrates the performance with varying sizes of hard-to-learn samples for refined unlabeled pool selection. 
With an acquisition size of 100 per iteration, the model achieves the optimal performance.
However, including any hard-to-learn samples in the sampling pool leads to a performance decline.
If the sample size is set too large, such as 200, the model initially shows relatively better performance due to the higher proportion of hard-to-learn samples in the early stages. 
However, as annotated data increases, performance declines because the hard-to-learn samples become less influential, necessitating a reduction in their selection. This parameter is adjustable across different datasets and model structures to ensure compatibility with the learning capabilities across varying scenarios. To deal with the problem of the amount of hard-to-learn samples decreasing with increased annotated data, CRAB adopts a decay function for the size of hard-to-learn samples to adapt to the training process. Figure~\ref{fig:decay} presents performance of three decay approaches, linear decay, cosine decay, and polynomial decay. Among these, polynomial decay achieves superior performance in terms of micro-F1 score, as it produces an accelerated decrease in output for sampling, better aligning with the trend in the size of hard-to-learn samples.


\subsection{Ablation study}
\begin{figure}
    \centering
    \begin{subfigure}{0.49\linewidth}
        \centering
        \includegraphics[width=\linewidth]{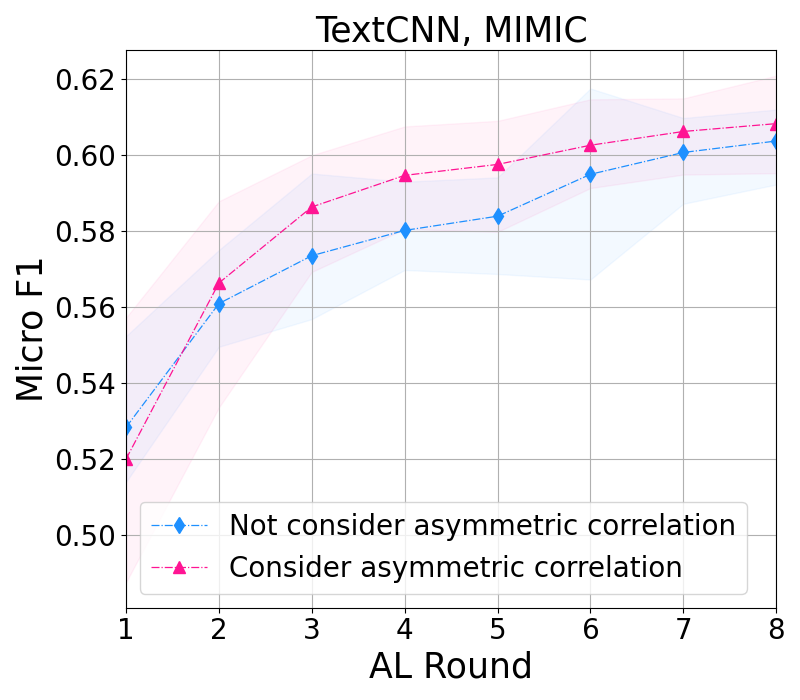}
        \caption{}
        \label{fig:hir}
    \end{subfigure}
    \hfill
    \begin{subfigure}{0.49\linewidth}
        \centering
        \includegraphics[width=\linewidth]{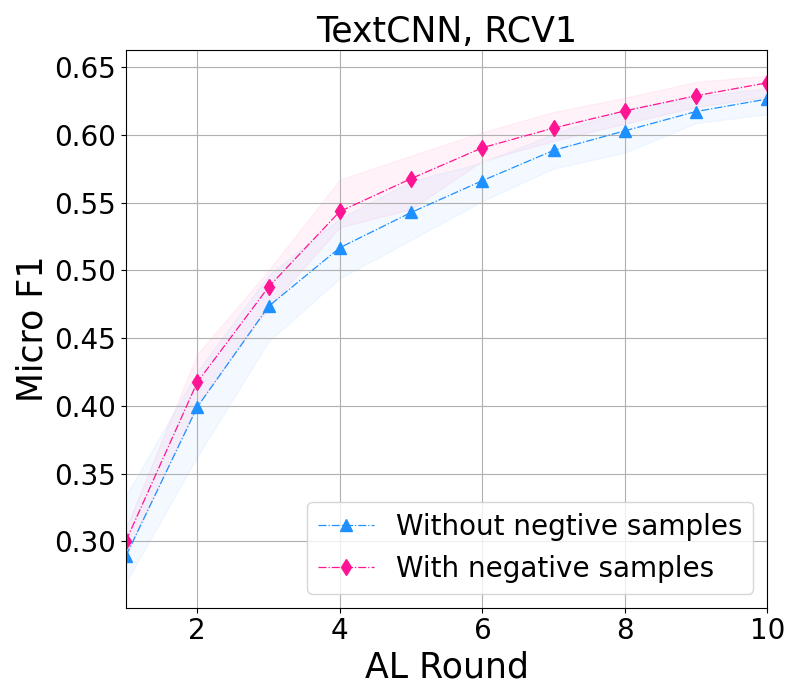}
        \caption{}
        \label{fig:noco}
    \end{subfigure}
    \caption{(a) Ablation study of the asymmetric-correlated label. (b) Ablation study of considering negatively correlated label pairs.}
    \label{fig:hir_noco}
\end{figure}

We conducted four experiments to examine whether the structure of CRAB improves MLAL performance by considering correlation. 
Taking into consideration the asymmetrically correlated label relationships, 
CRAB selects only the initial label in the correlation chain for per-label selection, 
thus avoiding duplicate selection of correlated labels. 
Figure~\ref{fig:hir} compares performance with and without considering asymmetrical correlations on the MIMIC dataset. 
Results illustrate that CRAB demonstrates superior performance 
and is more effective in querying indicative samples 
than when treating all labels equally. 
Figure~\ref{fig:noco} shows the benefits of sampling conflicted labels, 
with performance improvements becoming more pronounced in later training stages.

\begin{figure}
    \centering
    \begin{subfigure}{0.49\linewidth}
        \centering
        \includegraphics[width=\linewidth]{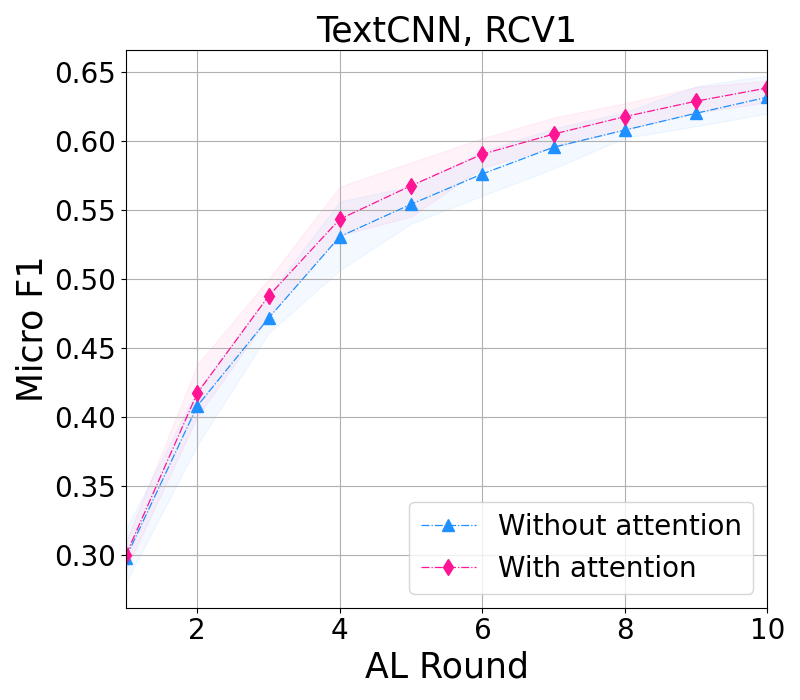}
        \caption{}
        \label{fig:attention}
    \end{subfigure}
    \hfill
    \begin{subfigure}{0.49\linewidth}
        \centering
        \includegraphics[width=\linewidth]{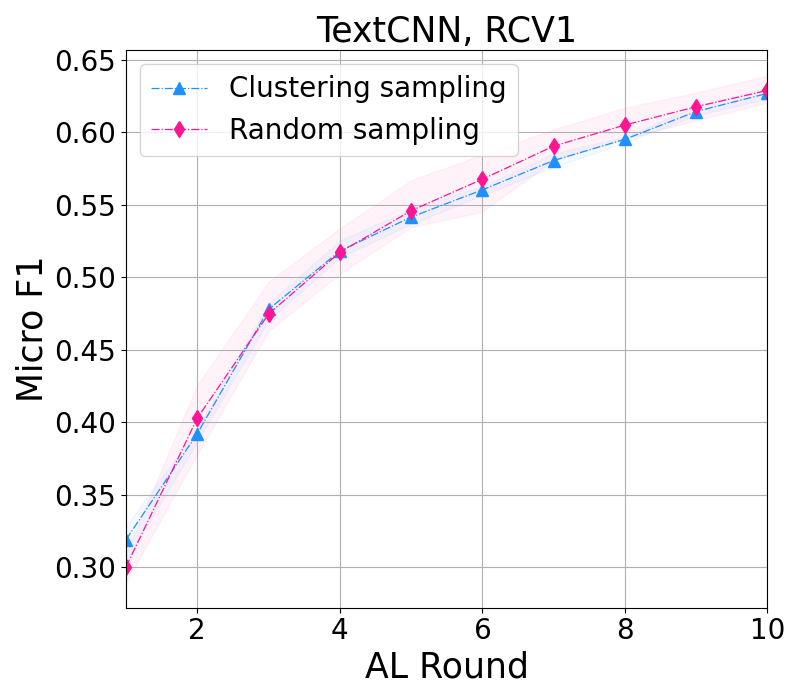}
        \caption{}
        \label{fig:sampling}
    \end{subfigure}
    \caption{(a) Ablation study of the correlation attention. (b) Performance for random sampling or cluster-based sampling for refined sampling pool selection.}
    \label{fig:att_random}
\end{figure}

Figure~\ref{fig:attention} illustrates how the correlation attention impact MLAL accuracy. 
To assess performance without correlation in score evaluation, 
we removed the correlation attention in Eq.~\eqref{eq11}, 
with the results shown by the blue line. 
Evidently, when positive label correlations are incorporated, 
CRAB performs more consistently throughout the experiment,
indicating that our strategy effectively models inter-label relationships to
make more informative queries.
Additionally, we compared the micro-F1 score and computation time of the random sampling and clustering-based sampling during the refined unlabeled pool sampling. 
As shown in Figure~\ref{fig:sampling}, 
random sampling performs almost identically to clustering-based sampling,
suggesting it can serve as a replacement during the refined unlabeled pool selection. 
Moreover, the querying time with random sampling decreases by 40\% . 
These findings demonstrate that our method is effective, efficient, and robust.


\section{Conclusion}
In this paper, 
we proposed an innovative MLAL query strategy, CRAB, which takes into account inherent label relationships within
a Bayesian framework. 
By updating the correlation matrices with the annotated data, our model is competent to query more representative samples in the initial stage and achieves a more accurate score for evaluating the informativeness of instances. 
Additionally, with the utilization of beta scoring rules, our model maintains consistently robust performance on imbalanced datasets. 
Leveraging pseudo labels and correlation-aware sampling, our strategy eliminates the need for additional training modules, and our model demonstrates significant performance improvements in MLAL on four benchmark datasets. 
Future research could explore the correlation at the instance space and investigate additional relationships between the data features and label distributions.

\begin{acknowledgements} 
We would like to thank the anonymous reviewers for their valuable and helpful comments. This research was supported by an Australian Government Research Training Program (RTP) Scholarship.
\end{acknowledgements}
\bibliography{ref}
\end{document}